\documentclass[letterpaper, 10 pt, conference]{ieeeconf}
\IEEEoverridecommandlockouts
\usepackage{cite}
\usepackage{amsmath,amssymb,amsfonts}
\usepackage[linesnumbered, ruled]{algorithm2e}
\usepackage[flushleft]{threeparttable}
\usepackage{siunitx} 
\usepackage{graphicx}
\usepackage{textcomp}
\usepackage{algpseudocode}
\usepackage{float}
\usepackage{subfig}
\usepackage{xcolor}
\usepackage{booktabs}

\def\BibTeX{{\rm B\kern-.05em{\sc i\kern-.025em b}\kern-.08em
    T\kern-.1667em\lower.7ex\hbox{E}\kern-.125emX}}

\title{Uncertainty-driven Exploration Strategies for Online Grasp Learning 
\thanks{$^{1}$Karlsruhe Institute of Technology, Karlsruhe, Germany. Email: yitian.shi@kit.edu. This work is completed during the author's affiliation with the University of Stuttgart, Stuttgart, Germany.}
\thanks{$^{2}$Bosch Center for Artificial Intelligence, Renningen, Germany. Email: firstname.lastname@de.bosch.com}
}

\author{Yitian Shi$^{1,2}$, 
        Philipp Schillinger$^{2}$,
		Miroslav Gabriel$^{2}$,
		Alexander Qualmann$^{2}$, \\
        Zohar Feldman$^{2}$,
		Hanna Ziesche$^{2}$,
		Ngo Anh Vien$^{2}$
 }

\begin{document}
\maketitle
\thispagestyle{empty}
\pagestyle{empty}

\begin{abstract}

Existing grasp prediction approaches are mostly based on offline learning, while, ignoring the exploratory grasp learning during online adaptation to new picking scenarios, i.e., objects that are unseen or out-of-domain (OOD), camera and bin settings, etc. In this paper, we present an uncertainty-based approach for online learning of grasp predictions for robotic bin picking. Specifically, the online learning algorithm with an effective exploration strategy can significantly improve its adaptation performance to unseen environment settings. To this end, we first propose to formulate online grasp learning as an RL problem that will allow us to adapt both grasp reward prediction and grasp poses. We propose various uncertainty estimation schemes based on \emph{Bayesian uncertainty quantification} and \emph{distributional ensembles}. We carry out evaluations on real-world bin picking scenes of varying difficulty. The objects in the bin have various challenging physical and perceptual characteristics that can be characterized by semi- or total transparency, and irregular or curved surfaces. The results of our experiments demonstrate a notable improvement of grasp performance in comparison to conventional online learning methods which incorporate only naive exploration strategies. Video: https://youtu.be/fPKOrjC2QrU
\end{abstract}

\section{Introduction}
Robotic picking or object grasping is a challenging task in robotics that requires the agent to select and execute the optimal grasp poses for various objects from observations of complex scenes, where the agent must search for an optimal grasping strategy in an exponentially large state space due to the inherent variability in object shapes, opaqueness, and materials, and in certain cases also due to the various characteristics of the robot's sensors and actuators. These challenges represent significant obstacles to designing hand-engineered algorithms as done in traditional approaches \cite{1309.2660}. To overcome those challenges, modern grasping methods have successfully applied advanced deep learning techniques that enable model-free grasp predictions for a wide variety of objects in unstructured environments \cite{kleeberger2020survey,newbury2023deep}. However, methods \cite{danielczuk2020exploratory,fu2022legs, gilles2022continual} are based on supervised learning and offline training, and can not generalize well to OOD objects or new environment settings.

To address these challenges, this paper focuses on tackling the problem of online grasp learning. It does so by leveraging exploration capabilities to systematically search the space of grasp configurations to find the best grasps for picking objects from an OOD bin. Those objects are characterized by unknown physical and perceptual features such as semi- or total transparency, and irregular or curved surfaces.

In particular, we propose a study with a realistic assumption that online grasp learning with exploration capabilities can be used to fine-tune or adapt a pretrained grasping network to a set of OOD objects. We focus on data-efficient offline-to-online learning \cite{2107.00591}, aiming to enable the agent to quickly learn from a limited amount of real-world interactions given a grasping network that was pretrained using small-scale offline learning. To this end, we propose to formulate online grasp learning as a reinforcement learning (RL) problem and to leverage the Convolutional Soft Actor-Critic (ConvSAC) algorithm \cite{2111.01510} to further update the policy. 
\begin{figure}
\hspace*{-.4cm}
  \centering
  \includegraphics[width=1.0\linewidth]{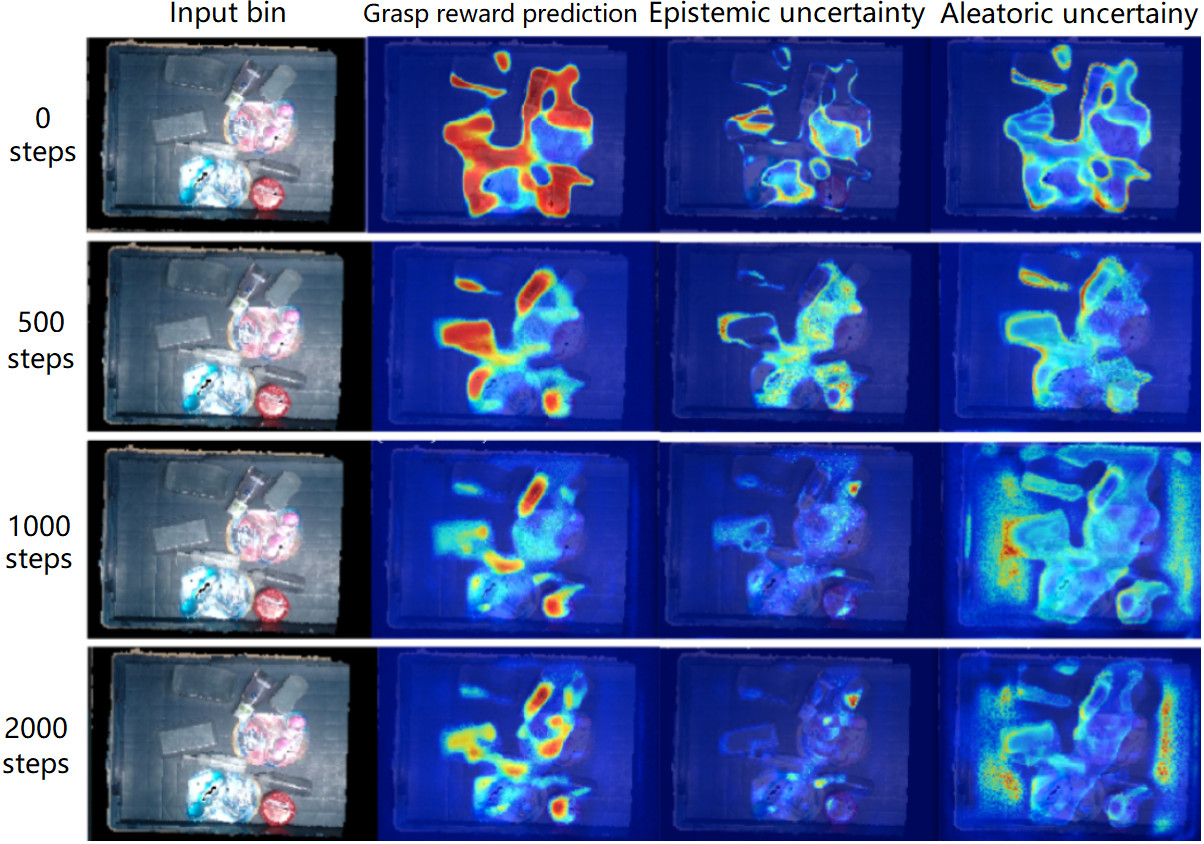}  
\hspace{-1\textwidth}\caption{\small Given an image of a picking scene (1st column), every 500 training steps our MV-ConvSACs predicts a grasp reward map (2nd column) and a normalized uncertainty map (3rd/4th columns).} 
\label{training_steps}
\end{figure}
We propose various exploration strategies that are based on uncertainty estimation to enhance the data efficiency and the domain generalization of the online learning process. To achieve this, we utilize Bayesian uncertainty estimation \cite{UDL} and quantile regression \cite{1710.10044,1905.09638} to compute a pixel-wise uncertainty map for a given input image. The uncertainty map allows the agent to actively choose the next picks that reduce ambiguity or uncertainty on the grasping scene while improving the grasp success in the long run, as depicted in Fig.~\ref{training_steps}. To summarize, our main contributions are as follows:
\begin{itemize}
\item The formulation of the exploration problem for online grasp learning allows us to integrate various exploration strategies into existing RL-based grasp learning methods to improve grasp performance online.
\item New architectures for uncertainty estimation and principled exploration strategies specifically for bin picking. These exploration strategies are based on pixel-wise uncertainty maps that can be computed using Bayesian uncertainty estimation or distributional regression.
\item Experiments and ablation studies on a real-world bin picking setup to i) demonstrate the proposed approach, and ii) understand the role of different types of uncertainties, i.e. epistemic and aleatoric.
\end{itemize}

\section{Related works}

\subsection{Modern Robot Grasping Methods}
Modern robot grasping methods often employ deep learning techniques trained on extensive datasets to predict grasps \cite{Dex, gilles2022metagraspnet}. Most strategies depend on generating a grasp reward map where each pixel represents the likelihood of a successful grasp at this location. For instance, Mahler et al. \cite{Dex} suggest the prediction of a grasp map for suction and parallel-jaw grasps through supervised datasets using RGB-D or depth as input. Morrison et al. \cite{MorrisonLC18} and Satish et al. \cite{satish2019policy} adopt this approach to predict both pixel-wise grasp reward maps and 4-DoF parallel-jaw grasp configurations.
Alternative approaches utilize point clouds \cite{yang2021robotic, jeng2020gdn, ni2020pointnet++, fang2020graspnet, li2020learning,sundermeyer2021contact} or voxels \cite{BreyerCOSN20} as input and predict dense grasp qualities and gripper configurations.
In a subsequent study, grasp prediction is trained jointly with object shape reconstruction \cite{0002ZSFZ21}.
Recent works introduce the concept of pixel-wise grasp maps and grasp configuration predictions for single-suction grippers \cite{cao2021suctionnet} and multi-suction cup grippers \cite{abs-2307-16488}.

\subsection{Deep RL-based Grasping Methods}

Vision-based RL has emerged as a promising approach to robotic grasping, which involves using visual information to guide robot actions through RL networks \cite{sur,1806.10293,levine2018learning} and are often end-to-end and optimize close-loop policies for grasp planning from raw visual inputs. One of the primary obstacles lies in the requirement of substantial quantities of training data of exceptional quality, owing to the high dimensionality of visual \cite{review} or depth \cite{8463204} input. Additionally, the algorithm must effectively extrapolate the acquired knowledge to unknown scenarios while maintaining its efficient performance. Open-loop RL methods are also used for 6-DoF bin picking scenarios \cite{1803.09956,BerscheidMK19,BerscheidFK21,2111.01510} which need a substantial number of updates to achieve ideal performance. However, these works often resort to training from scratch and use a standard exploration policy, i.e. a policy entropy bonus, hence they require an extensive amount of online samples.

Several works have proposed more principled exploration strategies for online grasp learning\cite{fu2022legs,danielczuk2020exploratory,lu2020multi,eppner2017visual,kroemer2010combining,li2020accelerating,laskey2015multi,oberlin2018autonomously}, but so far singulated object scenes were addressed and their extensions to bin picking is non-trivial. Alternatively, meta-learning or few-shot learning have also been applied for online learning for bin picking \cite{guo2022few,chen2022meta,barcellona2023fsg}. These works show that the learned grasp can be quickly adapted to OOD objects. However, they require a few shots of context grasps provided either by an oracle or by executing passive actions. In contrast, our approach gathers these context grasp points via an actively exploring policy.



\subsection{Uncertainty Estimation in Deep RL}
Uncertainty estimation is a research focus in various machine learning and deep learning subfields. Kendall et al. classify uncertainties into epistemic and aleatoric uncertainties\cite{1703.04977}. Epistemic uncertainty arises from the model's lack of knowledge or information. Aleatoric uncertainty, on the other hand, is related to the inherent stochasticity (or noise) in the data. In the RL domain, Charpentier et al. \cite{2206.01558} introduced desiderata influenced by supervised uncertainty estimation, encompassing both aleatoric and epistemic uncertainties. Lee et al. \cite{2007.04938} propose deep ensemble techniques for the Soft Actor-Critic (SAC) algorithm \cite{1801.01290}, utilizing uncertainty estimates to re-weight sample transitions during policy updates. By incorporating uncertainty measures, the agents gain insights into the reliability of the collected data. Clements et al. \cite{1905.09638} focused on estimating and disentangling uncertainties in distributional RL agents by estimating uncertainty at various quantiles \cite{1710.10044}, enabling more precise and reliable estimates of uncertainty in the RL setting. Additionally, methods based on Kalman filters as discussed in \cite{wagner2023kalman}, have also demonstrated their success in Bayesian uncertainty estimation.

\section{Uncertainty-driven Offline-to-Online Robotic Grasp Learning}
We consider the following online grasp learning problem setting.
Given an RGB-D image of the scene, our goal is to predict multi-channel maps for a suction gripper: a \emph{pixel-wise grasp reward map}, a \emph{pixel-wise grasp orientation (rotation) map} and a \emph{pixel-wise uncertainty (variance) map}.
These prediction maps can be used to derive an exploration grasping action. After each grasp, the network receives sparse reward feedback (success or failure) and is updated accordingly.
\begin{figure}
  \centering
  \includegraphics[width=1.02\linewidth]{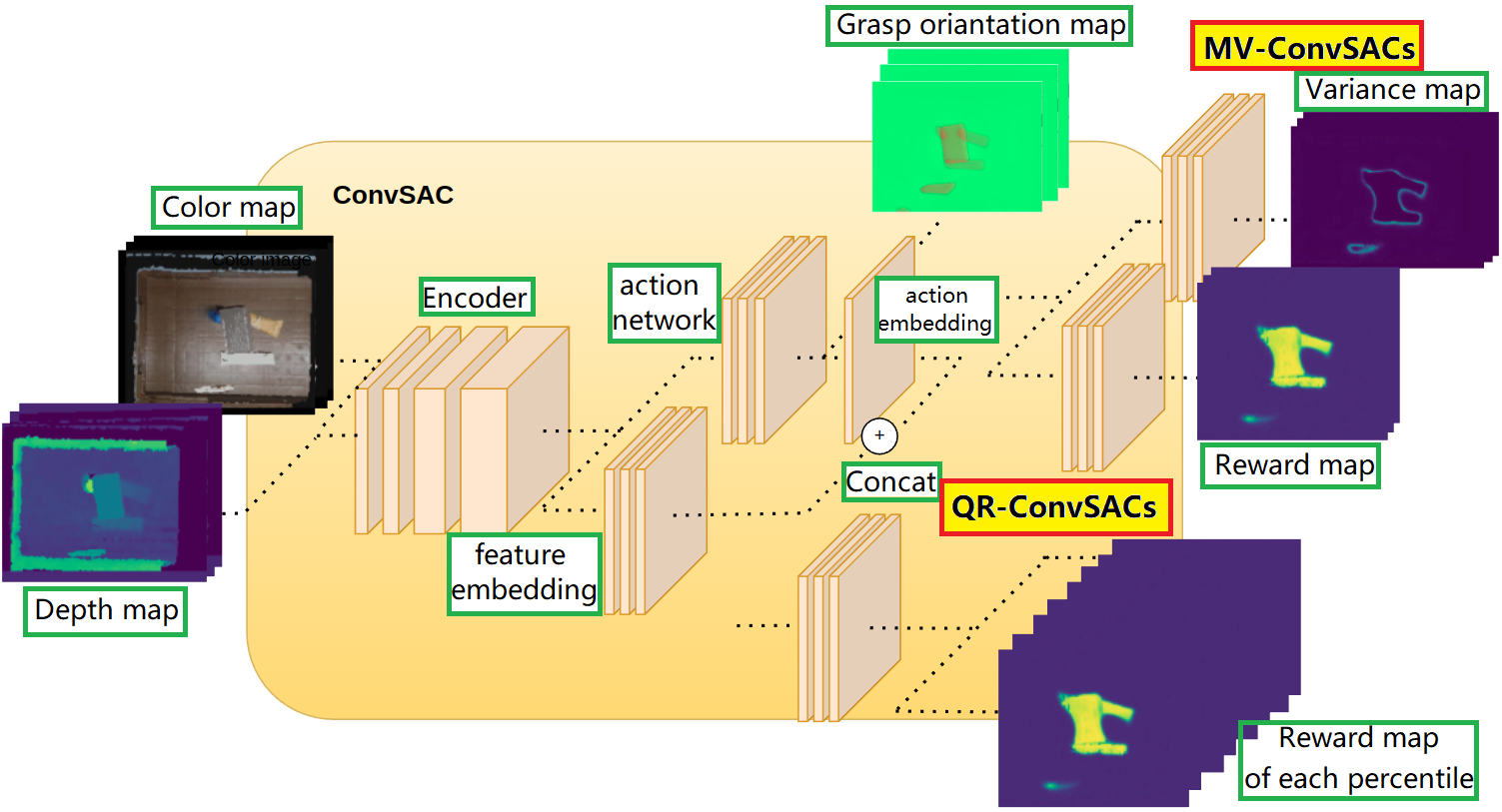}  
  \label{fig:sub-first}
  \caption{Ensemble of probabilistic ConvSAC network architecture: \textbf{MV-ConvSACs} and \textbf{QR-ConvSACs}.} 
\label{ConvSAC}
\end{figure}
\subsection{Problem Formulation}
\label{formulation}
Here we suppose to model the online grasp learning problem on a bin picking setting as an MDP $(\mathcal{S},\mathcal{A},\mathcal{T},r)$ with state space $\mathcal{S}$, action space $\mathcal{A}$, transition function $\mathcal{T}$ and reward function $r$. In each step, the system observes a state $s_t \in S$, an action $a_t \in \mathcal{A}$ is taken by the policy $\pi(a_t|s_t)$ and the reward is received from the environment $r(s_t, a_t)$. A new state $s_{t+1}$ follows upon based on the transition $\mathcal{T}$ from the environment. In the setting of \cite{2111.01510}, the state $s_t$ is represented by a set that is composed of one color image, normal map, and height map $s_t = (I_c, I_n, I_d)_t$ with $I_c \in \mathbb{R}^{H*W*3}$, $I_n \in \mathbb{R}^{H*W*3}$ and $I_d \in \mathbb{R}^{H*W*1}$. In our experiments, the states are captured by a stereo sensor with a top-down view of the object bin.
The action $a_t$ corresponds to a three-dimensional orientation represented by Euler angles $(\alpha_t,\beta_t,\gamma_t)$ and Cartesian coordinates $(x_t, y_t, z_t)$. Due to the rotational symmetry of our suction gripper ($\gamma_t$ ignored), we define the grasp action as $a_t=(x_t, y_t, \alpha_t,\beta_t) \in {\mathcal A}$ since $z_t$ is directly extracted from the height map. The reward $r_t$ is 1 when a successful grasp is executed or 0 otherwise. We aim to optimize a policy $\pi(s):{\mathcal S} \mapsto {\mathcal A} $ that maximizes the total grasp success return $\sum_t r_t$.

We follow a similar framework as proposed by ConvSAC \cite{2111.01510} and HACMan \cite{abs-2305-03942} to create network architectures for continuous actions, as well as $Q$-learning for discrete actions. In particular, the \textbf{Actor} module $\pi(s)$ infers pixel-wise Gaussian actions, resulting in an action map denoted as $A$. These actions are then concatenated with the embedding of the input image and evaluated by a \textbf{Critic} module, resulting in a $Q$-value map denoted as $Q$, namely \emph{grasp reward map}.

\subsection{Ensemble of Probabilistic ConvSAC}
Here we propose an extension of ConvSAC to model the critic's uncertainty using two strategies. The \emph{first strategy} is to model heteroscedastic aleatoric uncertainty following the Gaussian likelihood \cite{1703.04977} of each pixel of the Q-value map. The \emph{second strategy} is to predict the Q-value for each pre-defined distributional quantile \cite{1905.09638} individually. Figure \ref{ConvSAC} shows our proposed network architectures. The training of the actors is similar to the ConvSAC's actor training procedure, where we implement a probabilistic action as $\pi_\theta(s)\sim\mathcal{N}(A_{\mu}(s), A^2_{\sigma}(s))$.

\subsubsection{Gaussian-based Uncertainty Estimation}
Suppose we have the input data $s$, the network will encode it and generate the grasp orientation map $A_{\mu}(s) \in \mathbb{R}^{H*W*3} $ for each pixel $D = \{a_i(s)\}_{i<H*W, i\in \mathbb{N}}$. To criticize the pixel-wise action $a_{i}(s)$, the critic network will take the embedding from both the input feature as well as the action map and finally output the reward map or $Q$-value map as well as the variance map, denoted as $Q(s, A_{\mu}(s)) \in \mathbb{R}^{H*W*1}$ and $\mathrm{Var}(s, A_{\mu}(s)) \in \mathbb{R}^{H*W*1}$. We denote $i$ as the pixel index of a specific map for all the notations mentioned above.
For simplicity, we call this architecture as \textbf{Mean-Variance Convolutional Soft Actor-Critics (MV-ConvSACs)}. Specifically, we build another head of the ConvSAC's critic network to predict the variance map $\mathrm{Var}(s, A_{\mu}(s))$ parallel with the reward map $Q$. Inspired by \cite{2006.08903}, our critic network is trained in a supervised manner with negative log-likelihood (NLL) loss \cite{1703.04977}.

We further propose to improve the MV-ConvSACs framework using ensemble learning\cite{1612.01474}. Following \cite{2107.00591}, using multiple agents of Q-functions can achieve a higher resolution of pessimism for OOD data. 
Suppose $N$ MV-ConvSACs agents are considered with an ensemble of $Q$-functions and their variance $\{Q_j(s, A_{\mu}(s)), \mathrm{Var}_j(s, A_{\mu}(s))\}_{j=1}^N$. As a result, the epistemic and aleatoric uncertainties, and the total uncertainty can be computed as follows:
\begin{equation*}
\begin{aligned}
\bar{V}_{ale}(s, A_{\mu}(s))&= \frac{1}{N}\sum_{j=1} 
^N \mathrm{Var}_j(s, A_{\mu}(s))
\\
\bar{V}_{epi}(s, A_{\mu}(s))&= \frac{1}{N}\sum_{j=1} 
^N \Bigl(Q_j(s, A_{\mu}(s)) - \bar{Q} \Bigr)^2
\end{aligned}
\end{equation*}
\begin{equation*}
\bar{V}_{all}(s, A_{\mu}(s))= \bar{V}_{ale}(s, A_{\mu}(s)) + \bar{V}_{epi}(s, A_{\mu}(s))
\label{3.1}
\end{equation*}
where $\bar{Q} =\sum_j^N Q_j(s, \pi_{\mu}(s))$.

\subsubsection{Quantile-based Distributional Learning}
While \textbf{MV-ConvSACs} assume the critic's reward map $Q$ being a pixel-wise Gaussian distribution, quantile-based estimation allows us to capture the full range of possible outcomes and their associated probabilities. By using quantile estimations, we can easily estimate the value at a specific percentile of the reward distribution represented by individual percentiles. 

Here we aim to construct a discrete quantile-regressed critic estimator. Suppose the target reward map is a random variable $Z(s, \pi(s))\in \mathbb{R}^{H*W*1} $, we estimate the percentile corresponding to the $k$'th quantile $\tau_{k}$ by $Z_{\tau_{k}}(s, A(s))$ with $k\in [1, K]$ for $K$ quantiles in total. To realize this, we simply construct $K$ output heads for each quantile regression member. For notational simplicity, we denote the $k$'th quantile reward estimate $Z(s, A(s))$ as $\hat{Q}_{k}\in \mathbb{R}^{H*W*1}$. Given this architecture, the $Q$-value map is estimated as $Q(s, A(s))=\mathbb{E}_{k \sim \mathcal{U}[1,K]}[ \hat{Q}_{k} ]$. Similar to before,  we call this architecture as  \textbf{Quantile Regression Convolutional Soft Actor-Critics (QR-ConvSACs)}.

Inspired by this work \cite{jiang2022uncertainty} that proposes an application of ensemble learning for distributional Q-learning, we also use $N$ QR-ConvSACs agents with an ensemble of quantile heads $\{\hat{Q}_{k,j}(s,A(s)) \}_{j\in[1,N], k\in[1,K]}$. Given this, the estimate $Q$-value map is computed as:
\begin{align*}
    Q(s, A(s))=\frac{1}{K\times N}\sum_{k=1}^{K}\sum_{j=1}^{N} \hat{Q}_{k,j}(s,A(s))
\end{align*}    
The training of QR-ConvSACs' critics relies on quantile regression. In particular, following \cite{1905.09638} we propose to train the prediction heads according to the Huber loss \cite{1710.10044} respectively and model the estimated uncertainty as:
\begin{equation*}
\begin{aligned}
\bar{V}_{epi}(s)&=\frac{1}{K\times N}\sum_{k=1}^{K}\sum_{j=1}^{N}\left( \hat{Q}_{k,j}-\frac{1}{N}\sum_{j=1}^{N}\hat{Q}_{k,j}\right)^2
\\
\bar{V}_{ale}(s)&=\frac{1}{K}\sum_{k=1}^{K} \left( \frac{1}{N}\sum_{j=1}^{j} \hat{Q}_{k,j}-Q(s, A(s))\right)^2
\end{aligned}
\label{3.11}
\end{equation*}

\subsection{Data-efficient Online Training}

\subsubsection{Multi-processed Online Learning}
We build a pipeline for multi-processed online learning inspired by Ape-X \cite{1803.00933}, which is designed to efficiently collect valuable online data and train the agents in a scalable manner. We parallelize the training of individual agents from the ensemble on a single robotic picking cell. This distributed training pipeline achieves the ratio of {\bf 6:1} between \emph{the number of training steps vs. one online collected data sample} (one robot grasp), with an ensemble of $N=3$ agents. In addition, our pipeline can be theoretically extended to multi-robot data collection for better efficiency as well.

\subsubsection{UCB-based Exploration Policy}

Suppose in the online inference process, our stereo sensor captures the state $s$, and our ensemble agents give the correspondent reward map ${Q}(s, A_\mu(s))$, variance map ${V}(s, A_\mu(s))$ as well as the action map ${A}_\mu(s) = \frac{1}{N}\sum_{j=1}^N A_{\mu,j}(s)$ as the output action during inference. We design an exploration policy based using the UCB strategy \cite{1706.01502, wu2023uncertainty} as: 
\begin{equation}
{Q}_\mathrm{UCB}(s, A_\mu(s)) = {Q}(s, A_\mu(s)) + \delta \cdot {V}(s, A_\mu(s)),
\label{3.12}
\end{equation}
where $\delta\in \mathbb{R}_+$ is the UCB ratio, which is a hyperparameter that indicates the degree of exploration on uncertain pixel regions. In the experiment section, we will further investigate the effect of different uncertainty types, $\bar{V}_{ale}$, $\bar{V}_{epi}$ or the total uncertainty $\bar{V}_{all}$.

\subsubsection{Online Training with Data Buffer}

We are supposed to update our networks with the online collected data from the shared data buffer. Online learning utilizes similar objectives as offline training, while only on the selected pixel indexed by $i_{best}=\{h^{\ast},w^{\ast}\}$ that is selected according to the probabilistic policy in the inference process, where $(h^{\ast},w^{\ast})=\arg\max_{h',w'}Q_\mathrm{UCB}[h',w']$, and the best action parameter is extracted from the action map as $A[h^{\ast},w^{\ast}]$. Once the updates are completed, the new network parameters are sent to the parameter buffer periodically.

\section{Experiments}
\subsection{Experiment Setup}
Our study was conducted on the Franka Emika Robot equipped with a Schmalz suction gripper.
We aim to enable the robot to accurately detect and manipulate objects by mounting a Realsense d415 camera to capture a clear top-down view of the bin.
  \begin{figure}[htbp]%
    \centering
    \subfloat[Robot configuration]{
        \includegraphics[width=0.48\linewidth]{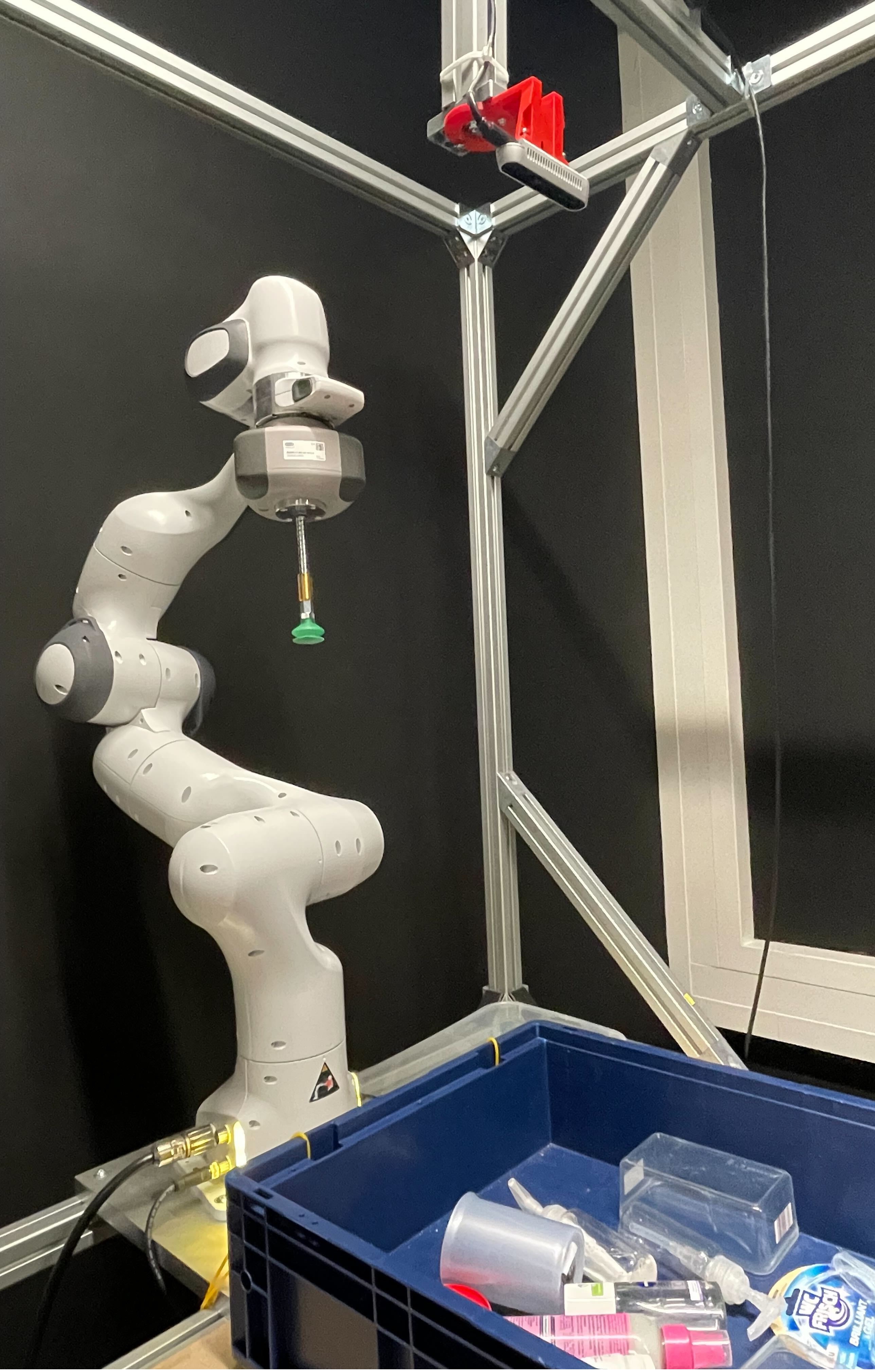}
        }\hfill
    \subfloat[Bin picking objects]{
        \label{rgb}\includegraphics[width=0.45\linewidth]{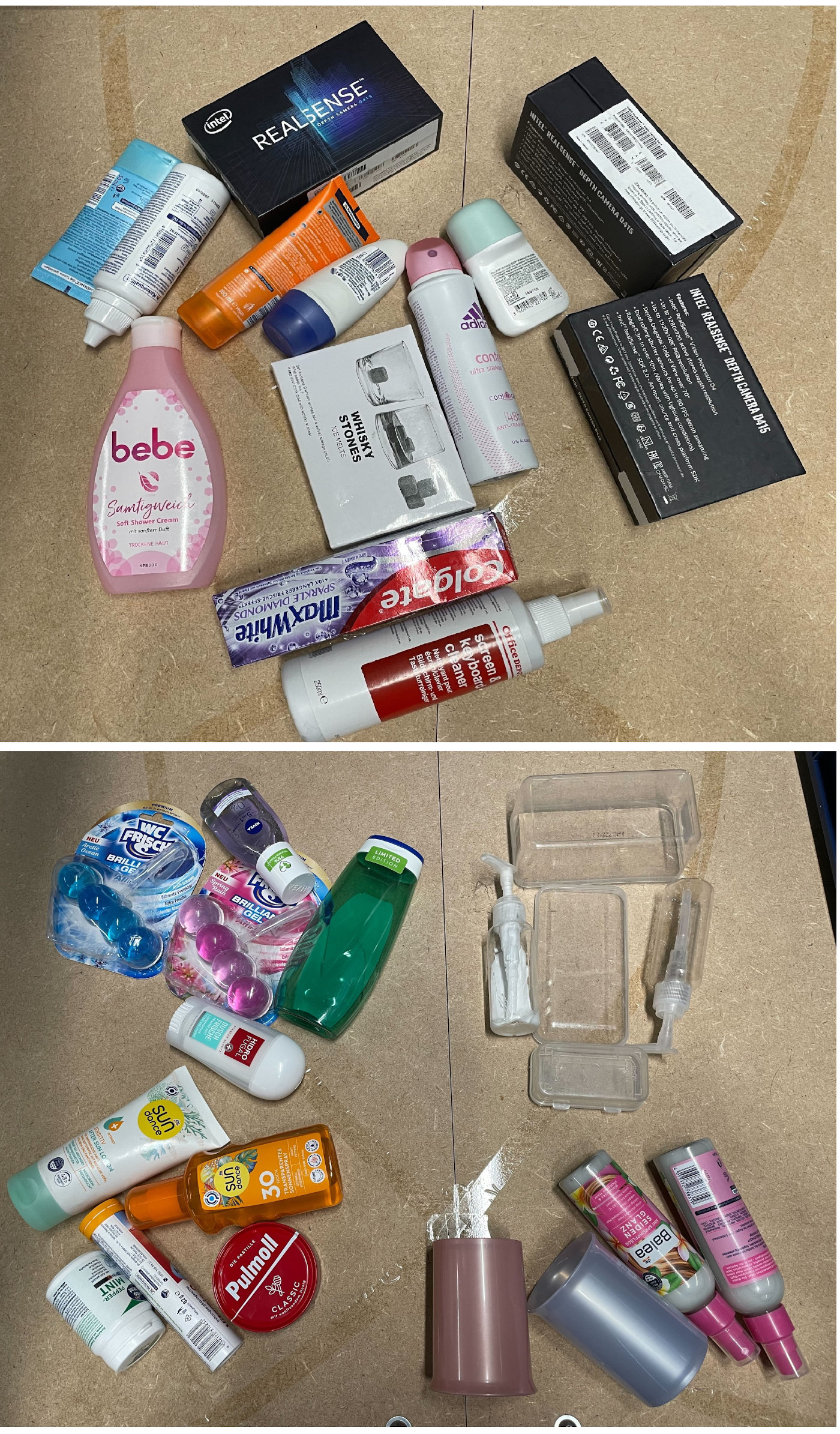}
\label{offline_object}
        }
        \\
    \caption{Grasping experiment setup}
            \label{robot1}
\end{figure} 
We prepared two sets of objects, each with specific characteristics and experimental purposes as in Fig. \ref{robot1} (b). The first object set contains opaque objects with rigid and regular shapes in Fig. \ref{robot1} (b) (top) including several rectangular boxes and cylindrical bottles with various sizes and textures. The second set contains "hard" objects with various characteristics that increase the difficulty of grasping in Fig. \ref{robot1} (b) (bottom) that are characterized by semi- or total transparency, irregular or curved surfaces, which are used to evaluate the performance of our proposed algorithm.

\begin{figure*}[htbp]%
    \centering
    \subfloat[Success rate (MV-ConvSACs)]{
    \centering
        \includegraphics[width=0.236\linewidth]{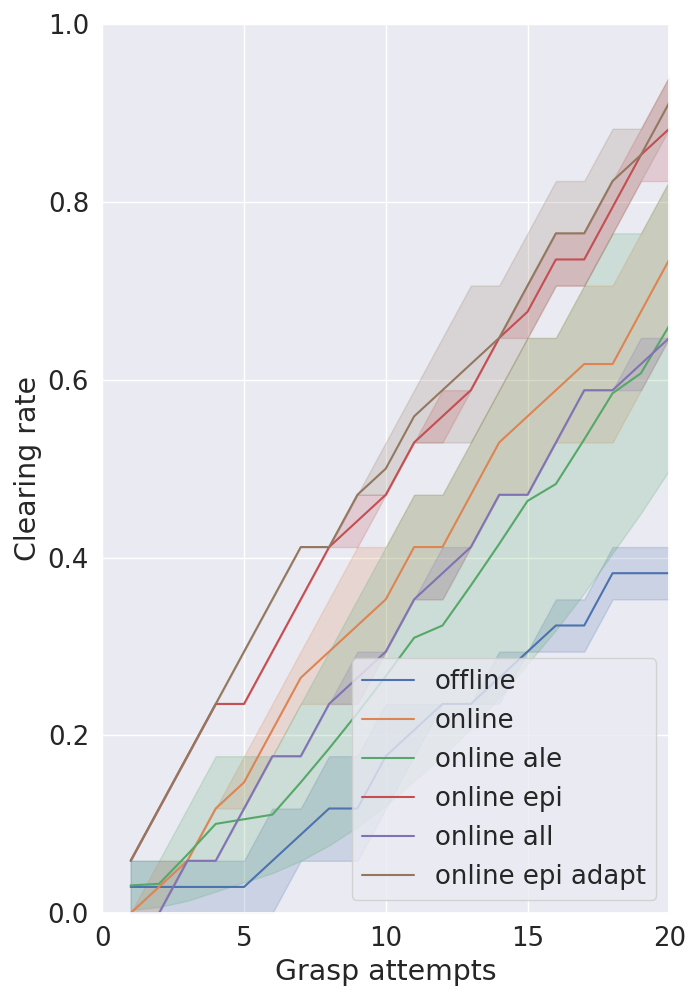}
        }\hfill
    \subfloat[Final success rate (MV-ConvSACs)]{
        \includegraphics[width=0.236\linewidth]{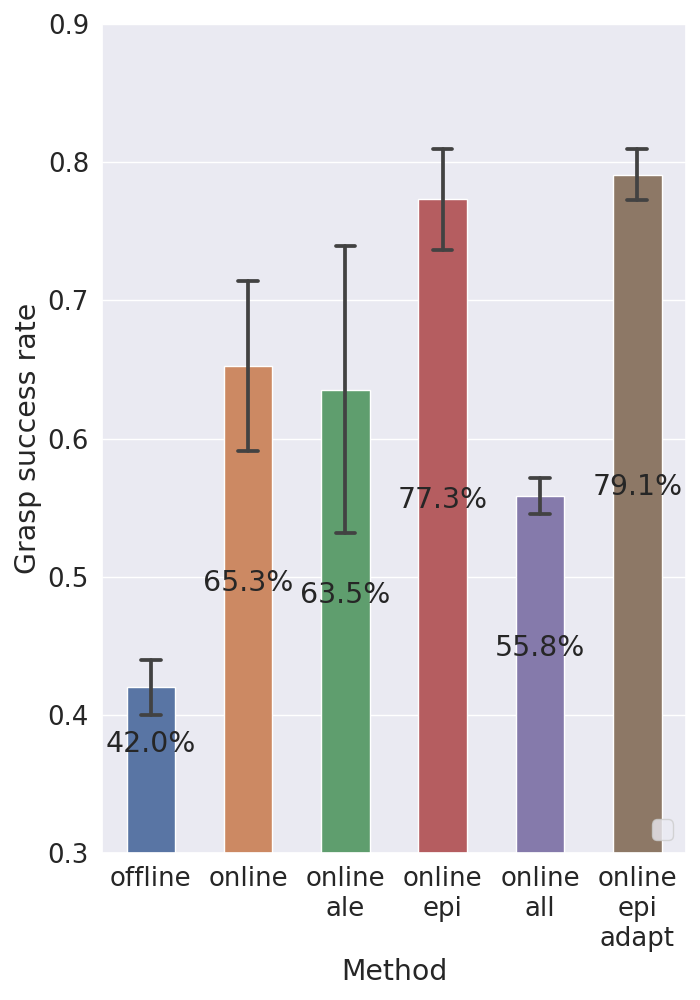}
        }\hfill
    \subfloat[Success rate (QR-ConvSACs)]{
        \includegraphics[width=0.236\linewidth]{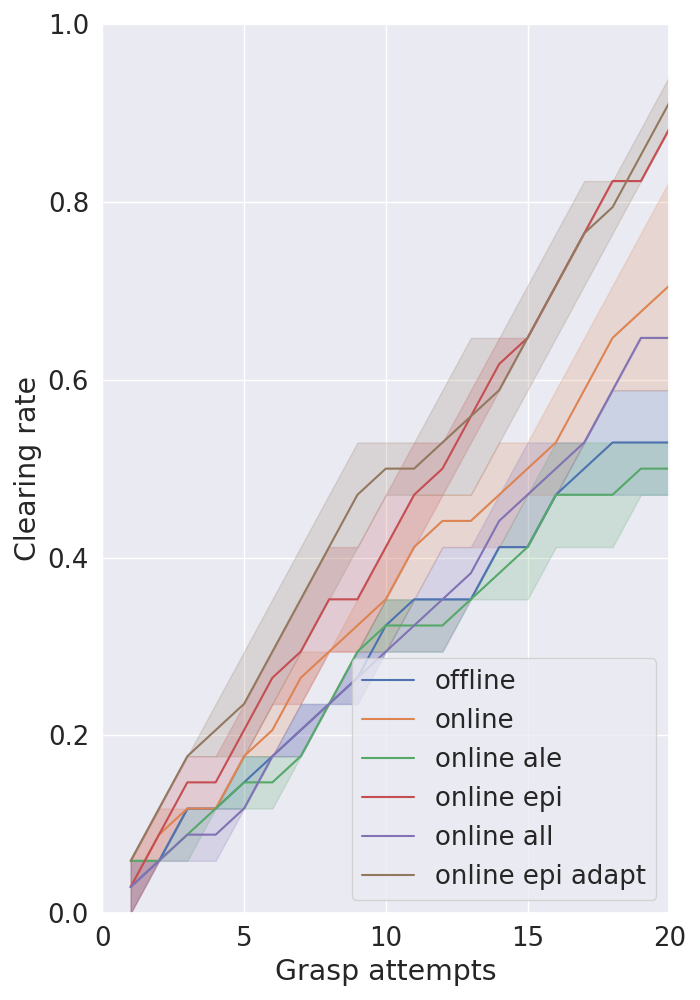}
        }\hfill
    \subfloat[Final success rate (QR-ConvSACs)]{
        \includegraphics[width=0.236\linewidth]{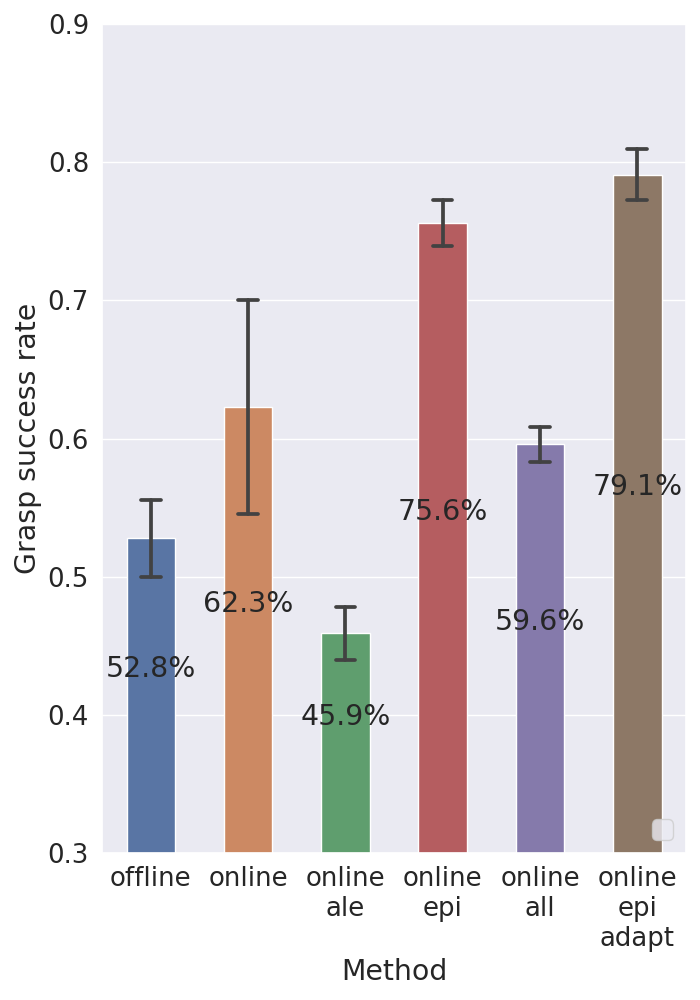}
        }
    
    \caption{Ablation on uncertainty exploration strategies for MV-ConvSACs and QR-ConvSACs}
\label{MV}
\end{figure*}
\begin{figure*}[htbp]%
    \centering
    \subfloat[Clearing rate (MV-ConvSACs)]{
        \includegraphics[width=0.236\linewidth]{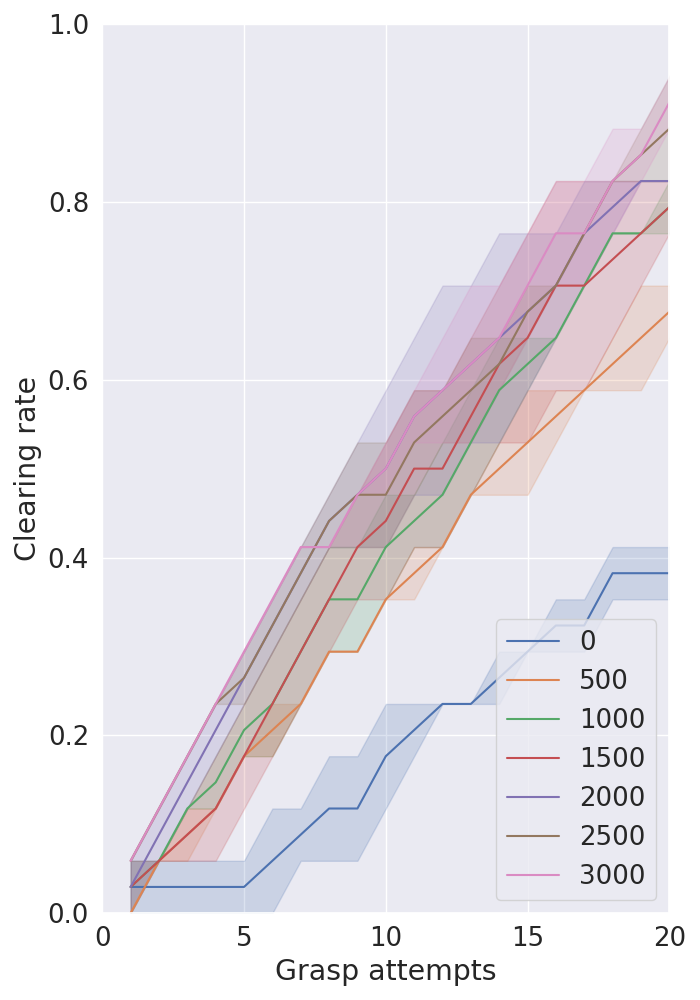}
        }
    \subfloat[Grasp success (MV-ConvSACs)]{
        \includegraphics[width=0.236\linewidth]{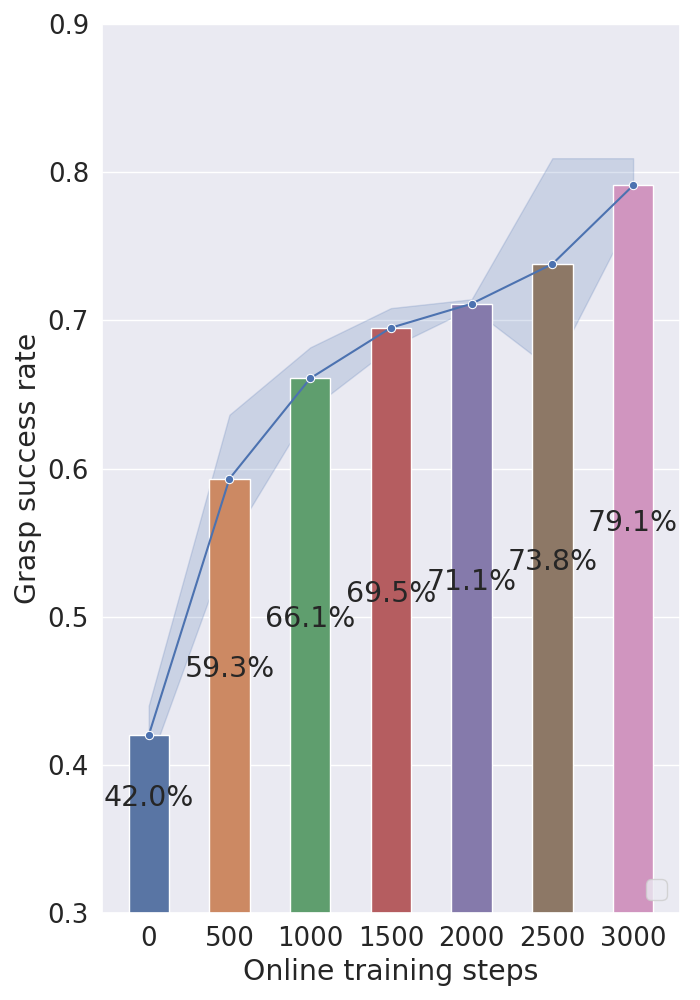}
        }
    \subfloat[Clearing rate (QR-ConvSACs)]{
        \includegraphics[width=0.236\linewidth]{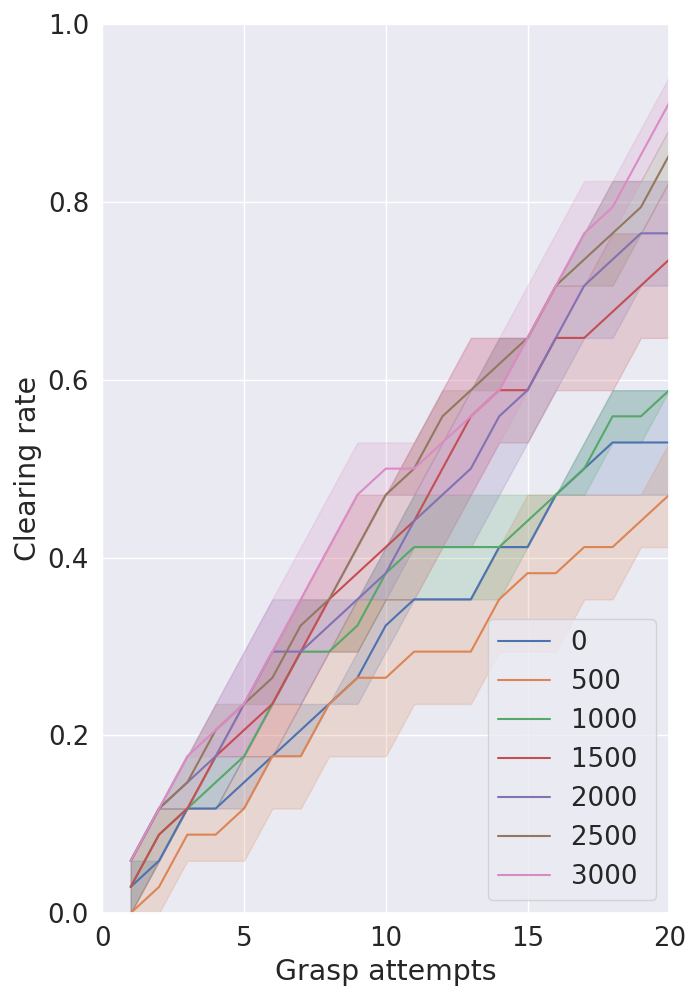}
        }
    \subfloat[Grasp success (QR-ConvSACs)]{
        \includegraphics[width=0.236\linewidth]{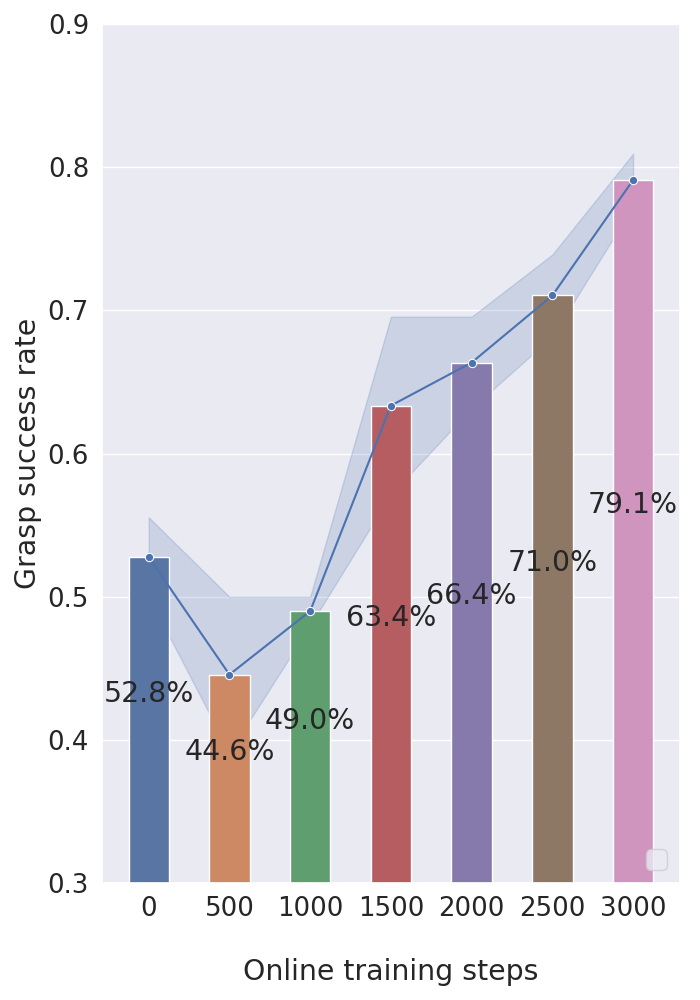}
        }
    
    \caption{Online training steps for MV- and QR-ConvSACs}
\label{abl}
\end{figure*}
\subsection{Technical Details}
Here we report the background configurations and hyper-parameter settings of our experiments. We use an ensemble of $N=3$ QR- or MV-ConvSAC agents. During both offline and online training, we trained both QR- or MV-ConvSAC and the baseline ConvSAC for 4000 steps with a mini-batch size of $12$ and a learning rate of $1e-4$ for both the actor and the critic. We apply data augmentation through proper affine (rotation and translation) and color jitter operations. For a fair comparison, all the models with different exploration strategies are trained for another 3000 steps in our ablation studies. During online learning, our robot will grasp from the bin with 10-17 randomly chosen objects from the online object set. We set the UCB ratio $\delta = 1$ in Eq. \eqref{3.12}. The period to transfer the networks from the training process to the inference process is set to be after every 10 steps of updates.
\paragraph{Offline training}
We first train all methods offline. The offline dataset consists of 300 scenes of random 5-10 objects. It can be represented by $\mathbb{D}_{offline}={\{I_c, I_n, I_d, A, Q\}_j}$, where $I_c, I_n, I_d$ are defined in Section \ref{formulation}, and $Q$ is the pixel-wise approximated ground truth $Q$-reward map. The approximated ground truth $Q$ map is computed simply using background subtraction to detect object regions, and then assign values proportional to the inverse of the standard deviation of the object surface normals \cite{abs-2307-16488}. The action map for a suction gripper at each pixel is assigned to its respective negative normal vector.
\paragraph{Baselines}
We evaluate and compare our methods to the two ensembles ConvSAC versions: offline-trained MV- and QR-ConvSAC (called \emph{offline} version, correspondingly), and these versions with online fine-tuning (called \emph{online} version) using exploration strategies from the standard SAC algorithm.
\paragraph{Metrics}
Our evaluation and comparison use two metrics: i) \emph{Grasp success rate}: The percentage of successful grasps over grasp attempts, and ii) \emph{Clearing rate}: The percentage of objects removed over the total objects in a bin.
\paragraph{Evaluation scenes} To achieve a fair comparison, we predefined a set of 2 evaluation scenes of 17 objects, which contains a selected subset from our online objects. We rearranged the scenes for each algorithm in a consistent manner, so that any observed differences in performance could be attributed solely to the algorithmic approach rather than to any differences in the scene arrangement.

\subsection{Online Learning Results}


\subsubsection{Ablation on the Online Exploration Strategies} In this section, we mainly focus on the uncertainty type for Gaussian-UCB exploration as in Eq. \eqref{3.12}. Figure \ref{MV} shows our results averaged over the evaluation scenes, where we evaluate the networks by setting an episode of 25 grasp attempts per scene. The episode can terminate earlier the all objects are cleared. The shading regions and error bars represent the mean estimate's first standard deviation. The results show that our online learning approaches MV- and QR-ConvSACs achieve a significant improvement over both ConvSAC ensemble baselines offline and online.

Figure \ref{MV} (a, b) shows that for MV-ConvSACs the exploration strategy using epistemic uncertainty gives the highest final clearing rate of above 90\%, while also showing a relatively high grasp success. QR-ConvSACs show a similar tendency, they also achieve comparable clearing rates and grasp success. Overall, epistemic uncertainty is shown to be particularly effective in guiding the agent's exploration towards promising actions and strategies, leading to more efficient and targeted experiences and resulting in faster and more effective learning. On the other hand, these results consistently prove that explorations using aleatoric uncertainty or total uncertainty can mislead the learning process, hence converge to a poorer policy than the baselines.


\begin{figure}[htbp]
    \subfloat[10 heads]{
        \includegraphics[width=0.32\linewidth]{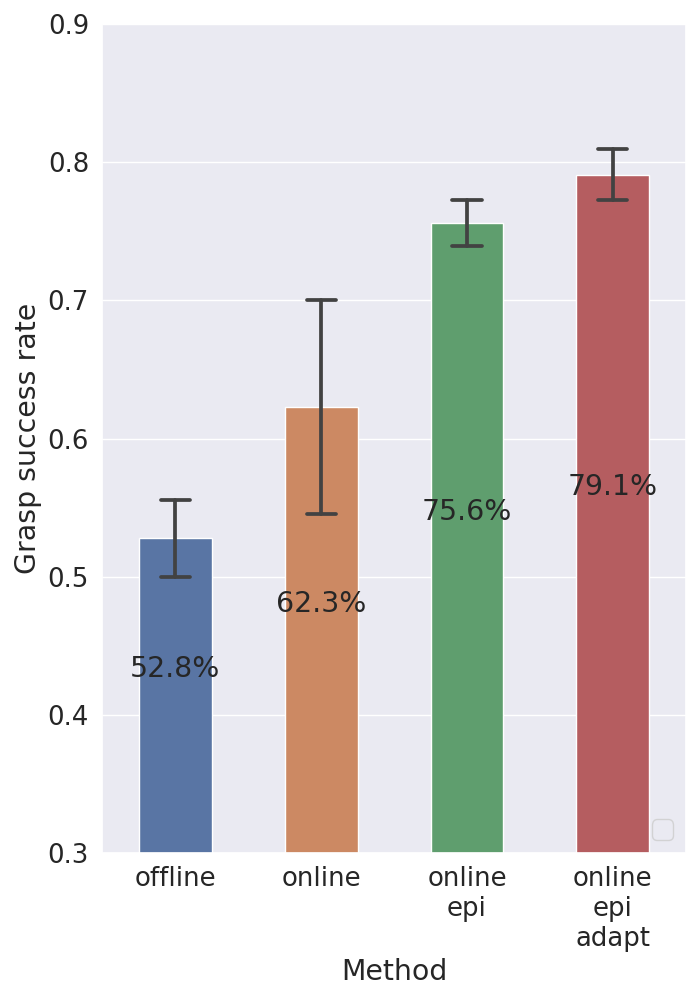}
        }
\hspace*{-0.4cm}
    \subfloat[20 heads]{
        \includegraphics[width=0.32\linewidth]{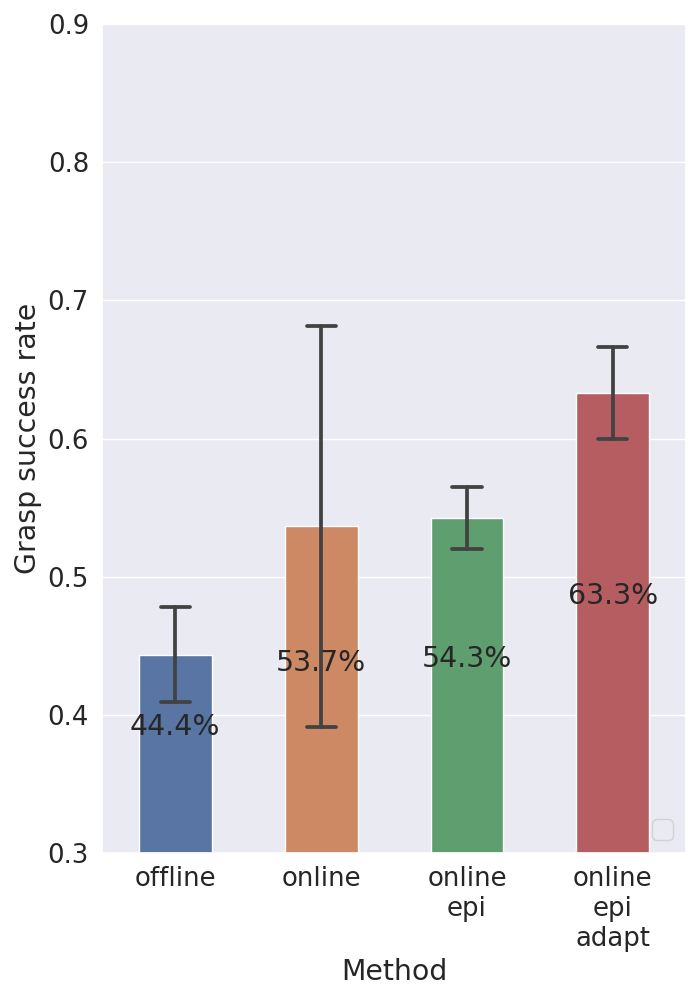}
        }
\hspace*{-0.4cm}
    \subfloat[100 heads]{
        \includegraphics[width=0.32\linewidth]{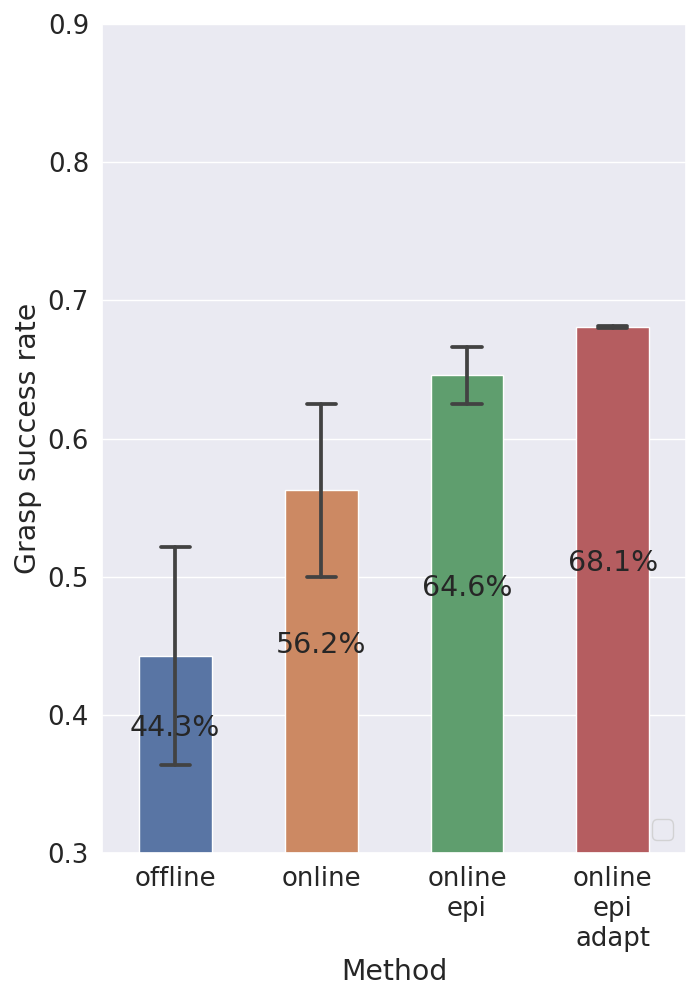}
        }
    \caption{QR-ConvSACs: Ablation of quantile head numbers}
\label{head}
\end{figure}
We further propose to adaptively tune the UCB ratio from 1 to 0 (marked as "online epi adaptive" in Fig. \ref{MV}) that follows a cosine decay. These results demonstrate that the adaptive exploration strategy performs better than the fixed UCB ratio in terms of clearing rates and grasp success. Specifically, we observe that a higher level of exploration is needed in the initial stage of learning, while a lower level of exploration is more appropriate in later stages to consolidate the learned knowledge.

\subsubsection{Ablation on the Number of Online Data}
We further conducted an ablation study on the amount of collected online data (proportional to the training steps) on the learning efficiency and consistency of our approaches. During the online learning of the exploration of epistemic uncertainty with adaptive UCB ratio, we recorded the checkpoints after every 500 training steps and evaluated on our evaluation scenes. The results are presented in Fig. \ref{abl}, which shows an overall increase in both metrics during online learning. 

To gain a deeper understanding of the performance of our framework, we visualize the evolution of reward maps and uncertainty maps during the training process, as shown in Fig. \ref{training_steps}. Our observations show that the high reward regions tend to concentrate on parts of objects that can be grasped safely with a high probability, such as the center region of a homogeneous surface. On the other hand, false positives on hard objects are identified as the main exploration targets during the online training. Exploring these areas can help the agent learn more effective grasping strategies in the presence of uncertainty. Similarly, transparent objects can lead to false negatives due to the difficulty of accurately sensing the object's shape and position. Exploring these regions can help the agent to improve its understanding of the environment and reduce the number of false negatives.

\subsubsection{Ablation on the Number of Quantile Heads of QR-ConvSACs}

In this final experiment in Fig. \ref{head}, we report our ablation study on the number of quantile heads of QR-ConvSACs, where the improvement of the success rate does not increase after 20. The final grasp success rate of adaptive epistemic exploration achieves the best performance if using the same number of heads, while the best performance is around 79 \% with 10 quantile heads. For each quantile head choice, the adaptive setting also has the highest clearing rates, with above 90\%, 70\%, and 75\% for 10, 20, and 100 quantiles, respectively.
 

\section{Conclusions}
In this work, we have proposed uncertainty-based exploration for online grasp learning. The proposed strategies are based on deep ensemble learning, uncertainty estimation, and UCB exploration. Specifically, we investigate and employ two ways to integrate uncertainty estimation, including mean-variance estimation (MV-ConvSACs) and quantile regression (QR-ConvSACs). During training, we online train the offline initialized agents on real-world bin picking scenes. 
Based on a view that integrating uncertainty estimation into the algorithm improved its performance by providing more potential regions to be explored, we studied different exploration strategies based on the UCB algorithm by utilizing different types of uncertainties and found that epistemic uncertainty is much more informative than aleatoric uncertainty. Epistemic uncertainty refers to uncertainty due to a lack of knowledge, which can be completed by further grasp trials. While aleatoric uncertainty is due to randomness or noise in the data that is generally unavoidable. To sum up, our proposed methods show a promising overall success rate, generalization ability, and efficiency in the bin picking of unseen objects. Our future work will look at an extension into online learning of non-prehensile manipulation. In addition, improved exploration strategies on a fused space between visual and proprioceptive modalities will open more applications to other manipulation tasks. 

\bibliographystyle{IEEEtran}
\bibliography{bibfile}



\end{document}